\documentclass{article}

\usepackage{microtype}
\usepackage{graphicx}
\usepackage{subfigure}
\usepackage{booktabs} 

\usepackage{hyperref}

\usepackage[accepted]{icml2025}

\usepackage{amsmath}
\usepackage{amssymb}
\usepackage{mathtools}
\usepackage{amsthm}

\usepackage{url}
\usepackage{wrapfig}
\usepackage{comment}
\usepackage{orcidlink}
\usepackage{marvosym}

\usepackage[capitalize,noabbrev]{cleveref}

\theoremstyle{plain}

\theoremstyle{definition}

\theoremstyle{remark}

\usepackage[textsize=tiny]{todonotes}

\icmltitlerunning{Evaluating Cumulative Spectral Gradient as a  Complexity Measure}

\begin{document}

\twocolumn[
\icmltitle{Evaluating Cumulative Spectral Gradient as a Complexity Measure}




\icmlsetsymbol{equal}{*}

\begin{icmlauthorlist}
\icmlauthor{Haji Gul \orcidlink{0000-0002-2227-6564}}{udb}
\icmlauthor{Abdul Ghani Naim \orcidlink{0000-0002-7778-4961}}{udb}
\icmlauthor{Ajaz Ahmad Bhat\textsuperscript{\Letter} \orcidlink{0000-0002-6992-8224}}{udb}\\
\end{icmlauthorlist}


\icmlaffiliation{udb}{School of Digital Science, Universiti Brunei Darussalam, Jalan Tungku Link, Gadong BE1410, Brunei Darussalam}


\icmlcorrespondingauthor{Ajaz Ahmad Bhat}{ajaz.bhat@ubd.edu.bn} 


\icmlkeywords{Machine Learning, ICML}

\vskip 0.3in
]



\printAffiliationsAndNotice{\icmlEqualContribution} 

\begin{abstract}

Accurate estimation of dataset complexity is crucial for evaluating and comparing link‐prediction models for knowledge graphs (KGs). The Cumulative Spectral Gradient (CSG) metric \cite{branchaud2019spectral} —derived from probabilistic divergence between classes within a spectral clustering framework— was proposed as a dataset complexity measure that (1) naturally scales with the number of classes and (2) correlates strongly with downstream classification performance. In this work, we rigorously assess CSG’s behavior on standard knowledge‐graph link‐prediction benchmarks—a multi‐class tail‐prediction task— using two key parameters governing its computation: $M$, the number of Monte Carlo–sampled points per class, and $K$, the number of nearest neighbors in the embedding space. Contrary to the original claims, we find that (1) CSG is highly sensitive to the choice of $K$, thereby does not inherently scale with the number of target classes, and (2) CSG values exhibit weak or no correlation with established performance metrics such as mean reciprocal rank (MRR). Through experiments on FB15k‐237, WN18RR, and other standard datasets, we demonstrate that CSG’s purported stability and generalization‐predictive power break down in link‐prediction settings. Our results highlight the need for more robust, classifier-agnostic complexity measures in  KG link-prediction evaluation.
\end{abstract}
\section{Introduction}
\label{intro}

Knowledge graphs (KGs) underlie many high‐impact applications—ranging from recommendation systems ~\cite{spillo2024evaluating} and question answering~\cite{zeng2025kosel} to drug discovery ~\cite{zhang2025comprehensive,gul2025mucos}. By encoding relational knowledge as triples $(h,r,t)$, link prediction $(h,?,t)$ and entity prediction $(h,r,?)$ tasks on KGs enable models to infer missing relations or tail entities ~\cite{gul2024contextualized,gul2025muco}. These benchmarks however, remain challenging due to imbalanced class distributions and overlapping feature patterns across relations and entities ~\cite{bourli2020bias}. While metrics like MRR and Hits@k evaluate how accurately models retrieve correct links, these do not provide us a direct measure of the intrinsic complexity of KG datasets under various link prediction scenarios. A robust class‐separability measure would (a) quantify dataset complexity across different link‐prediction formulations—revealing, for example, whether predicting rare drug‐target pairs is inherently harder than predicting common entity relations— (b) anticipate generalization performance—setting realistic expectations for new methods before expensive downstream evaluation, and (c) facilitate a unified estimate of model performance across datasets.


CSG is a recently proposed spectral metric—derived from the eigenvalues of the normalized graph Laplacian—designed to quantify dataset complexity by measuring class separability. In image classification benchmarks, higher CSG values correlate strongly with lower test accuracy \cite{branchaud2019spectral}. However, KG link‐prediction classification is a large‐scale, multi‐class task with thousands of candidate tails (every graph entity). In this regime, two core claims of CSG merit re‐evaluation:
\begin{itemize}
    \item CSG’s reliance on the nearest‐neighbor parameter $K$ may prevent it from naturally scaling when the number of target classes grows to typical KG sizes.
    \item Although CSG correlates with accuracy in image tasks, it is unknown whether CSG scores—computed over embeddings from KG models (e.g., BERT‐based or translational)—correlate with standard KG metrics like Mean Reciprocal Rank (MRR).
\end{itemize}
No prior work has systematically evaluated CSG on canonical KG benchmarks (e.g., FB15k-237, WN18RR) or examined its sensitivity to the Monte Carlo sample size $M$ and neighbor count $K$. Empirical scrutiny is therefore required to assess CSG’s stability and predictive power in large‐scale link‐prediction settings.

To this end, we conduct the \textbf{first systematic evaluation of CSG in multi‐class tail‐prediction tasks across multiple standard KG datasets} (e.g., FB15k‐237, WN18RR). Specifically for each head–relation pair, we treat every candidate tail entity as a separate class. We vary the Monte Carlo sample size \(M\) and the nearest‐neighbor count \(K\) to compute and relate CSG values over embeddings. Further on, we compare CSG values against actual link‐prediction performance (MRR) to quantify how well CSG predicts generalization. We list below key findings of our work:
\begin{enumerate}
  \item \textbf{Sensitivity to \(K\):} CSG values change dramatically as \(K\) varies, showing that any perceived “scalability” with the number of classes is an artifact of specific dataset choices rather than an inherent property of CSG.
  \item \textbf{Weak Performance Correlation:} Across all datasets and KG models, CSG scores exhibit near‐zero Pearson correlation with MRR, contradicting the claim that CSG reliably predicts downstream accuracy.
\end{enumerate}

These results question CSG’s utility as a model‐agnostic separability metric for large‐scale classification and highlight the need for more robust measures in KG evaluation.

\section{Methodology}
\begin{figure}[ht]
\begin{center}
\centerline{\includegraphics[width=1.1\columnwidth]{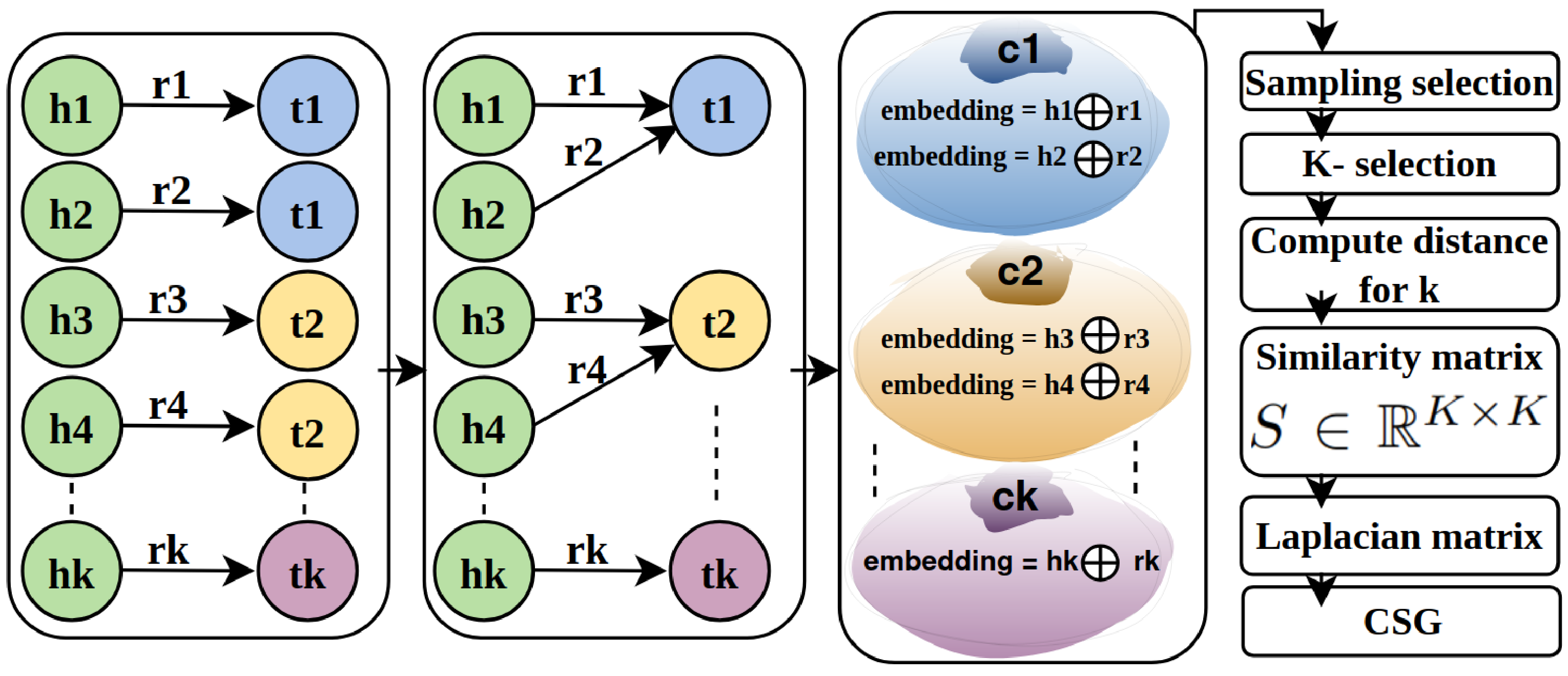}} 
\caption{\scriptsize Left box showing triplets where the heads are \((h_1, h_2, \ldots, h_k)\) green, relations \((r_1, r_2, \ldots, r_k)\), tails \((t_1, t_2, t_3)\) are in blue, yellow and purple. The next box denotes the grouping of their tail entities into classes: \(c_1\) for \(t_1\), with \((h_1, r_1, t_1)\) and, \((h_2, r_2, t_1)\) belonging to the same class, for example. BERT is used to embed head-relation pairs, producing 768-dimensional vectors, and then concatenates them, such as \(h_1 \oplus r_1\) and \(h_2 \oplus r_2\), for class \(c_i\). Next, a sampled \textit{K} search is performed to compute distances and a similarity matrix \(S \in \mathbb{R}^{K \times K}\). The Laplacian matrix \(L\) is obtained, and \(S\) the spectral complexity of the KG is quantified using the CSG calculated from its eigenvalues.}
\label{m-csg}
\end{center}
\end{figure}

 In our approach for CSG computation, we transform KG triplets into multi-class representations, use BERT embeddings for semantic richness, and apply spectral analysis to derive the CSG values; see Figure \ref{m-csg} for more clarification. The following subsections detail each step of this process.

\textbf{\textit{Grouping by Tail Entities:}} Knowledge graphs, such as FB15k-237 and WN18RR, consist of a set of triplets:
\begin{equation}
T = \{(h_i, r_i, t_i) \mid h_i \in E, r_i \in R, t_i \in E \},
\end{equation}
where \( h_i \) is the head entity, \( r_i \) is the relation, and \( t_i \) is the tail entity, with \( E \) being the set of all entities and \( R \) the set of all relations. The next step organizes this data by grouping triplets according to their tail entities class \(C\) (\textit{each unique} \( t_i \rightarrow \) \textit{denotes a unique class} \( C_i \)) using a mapping function:
\begin{equation}
G(C_i) = \{ (h, r) \mid (h, r, C_i) \in T \}, \quad \forall C_i \in E,
\end{equation}
resulting in a mapping:
\begin{equation}
C_i \mapsto G(C_i),
\end{equation}
which aggregates all \( (h, r) \) pairs pointing to the same tail \( C_i \). Each unique tail entity is treated as a distinct class, forming a set:
\begin{equation}
C_i = \{C_1, C_2, \ldots, C_K\},
\end{equation}
\( K \) is the total count of unique tails, generating \( K \) classes based on tail entities for further examination. 

\textbf{Generating Embeddings:}  
To transform textual head entities and relations into numerical form, a pre-trained BERT model generates dense vector embeddings. For each head entity \( h \) and relation \( r \), embeddings are:
\begin{equation}
e_h = \text{BERT}(h) \in \mathbb{R}^d, \quad e_r = \text{BERT}(r) \in \mathbb{R}^d,
\end{equation}
\( d \) the embedding dimension is. BERT-Base (Hugging Face Transformers) was used to generate 768-dimensional embeddings, preprocessing head entities and relations as single tokens. For every triplet \( (h, r, C_i) \in T \), a composite vector is formed:
\begin{equation}
\phi(h, r) = e_h \oplus e_r \in \mathbb{R}^{2d},
\end{equation}
where \( \oplus \) denotes concatenation. These composite vectors are then grouped according to their corresponding tail entities:
\begin{equation}
\Phi(C_i) = \{ \phi(h, r) \mid (h, r, C_i) \in T \},
\end{equation}
each tail \( C_i \) is associated with a set of \( (h, r) \) vectors.
This step provides a meaningful representation of the triplet data, organized by tail classes, preparing the data for complexity analysis.
\vspace{-3pt}

\textbf{Similarity Computation and Matrix Construction:}  
A similarity matrix \( S \in \mathbb{R}^{K \times K} \) is constructed, where \( K \) is the number of classes. Let \( \Phi(C_i) \) denote the set of vectors for class \( C_i \). Each vector:
\begin{equation}
\phi_m = e_h \oplus e_r \in \mathbb{R}^{2d}.
\end{equation}
A subset is sampled:
\begin{equation}
\begin{split}
M &= \min(N, |\Phi(C_i)|), \\
\Phi(C_i)_{\text{sample}} &= \{\phi_1, \phi_2, \ldots, \phi_M\} \subset \Phi(C_i)
\end{split}
\end{equation}
where \( N \) is the number of vector samples per class. 
To manage computational complexity for large KGs, we sample $M = 120$ vectors per class.
For each \( \phi_m \in \Phi(C_i)_{\text{sample}} \), compute its \( k = 50 \)-nearest neighbors via L2 distance:
\begin{equation}
    \|\phi_m - \phi_n\|_2^2 = \sum_{l=1}^{2d} (\phi_{m,l} - \phi_{n,l})^2.
\end{equation}
it computes the Euclidean distance between two concatenated BERT embeddings, 
\( \boldsymbol{\phi}_m \) and \( \boldsymbol{\phi}_n \), where \( \boldsymbol{\phi}_m, \boldsymbol{\phi}_n \in \mathbb{R}^{2d} \) represent the combined head-relation embeddings of triplets, respectively. while \( \phi_{m,l} \) and \( \phi_{n,l} \) denote the \( l \)-th components of the respective vectors. This distance metric is used during \( k \) values neighbor (k-NN) search to measure nearest neighbor triplets grouped by tail entities, enabling the construction of the class similarity matrix \( S \), which can be defined as in Equation \ref{eq:sim}. The distance computation directly impacts the spectral analysis by indicating how tightly or loosely classes overlap, thereby influencing the Cumulative Spectral Gradient (CSG), a measure of dataset complexity derived from the eigenvalue gaps in the graph Laplacian.
\begin{equation}
\label{eq:sim}
    S_{ij} = \frac{1}{Mk} \sum_{\phi_m \in \Phi(C_i)_{\text{sample}}} \sum_{\phi_n \in K(\phi_m)} \mathbb{I}[\phi_n \in \Phi(C_j)],
\end{equation}
where the indicator function is:
\begin{equation}
    \mathbb{I}[\phi_n \in \Phi(C_j)] = \begin{cases}
1, & \text{if } \phi_n \in \Phi(C_j), \\
0, & \text{otherwise}.
\end{cases}
\end{equation}
$\phi_n$ is an embedding vector, $\Phi(C_j)$ denotes the set of embeddings for class $C_j$, and $I$ is an indicator function returning $1$ if $\phi_n$ belongs to $C_j$. It is employed in the formation of the similarity matrix $S$ to enumerate the $K$-nearest neighbors of the class $C_i$ that belong to $C_j$, quantifying inter-class overlap for complexity analysis.

\textbf{Graph Laplacian and Spectral Analysis:} 
Graph Laplacian captures the connectivity and clustering tendencies of the classes, rooted in graph theory and spectral analysis. The normalized Laplacian provides a standardized measure of how classes are interconnected, accounting for variations in their degrees of connection. The graph Laplacian captures class connectivity and clustering tendencies. The diagonal degree matrix \( D \in \mathbb{R}^{K \times K} \) can be defined as Equation \ref{eq:dii}, while the normalized Laplacian as Equation \ref{eq:l}.
\begin{equation}
\label{eq:dii}
    D_{ii} = \sum_{j=1}^{K} S_{ij}, \quad D_{ij} = 0 \text{ for } i \ne j.
\end{equation}
\begin{equation}
\label{eq:l}
    L = I - D^{-1/2} S D^{-1/2},
\end{equation}
where \( I \) is the \( K \times K \) identity matrix, and:
\begin{equation}
\label{eq:dii12}
    D_{ii}^{-1/2} = \frac{1}{\sqrt{D_{ii}}}, \quad \text{for } D_{ii} > 0.
\end{equation}
$D_{ii} = \sum_{j=1}^{K} S_{ij}$ representing the total similarity of a class $C_i$ to all other classes, where $D_{ii}^{-1/2} = \frac{1}{\sqrt{D_{ii}}}$, ensures eigenvalues. Compute eigenvalues \( \lambda_0, \lambda_1, \ldots, \lambda_{K-1} \) and eigenvectors \( u_1, u_2, \ldots, u_K \) from:
\begin{equation}
\label{eq:lui}
    L u_i = \lambda_i u_i, \quad u_i \in \mathbb{R}^K, \quad \|u_i\| = 1, \quad 0 \le \lambda_i \le 2.
\end{equation}
yields eigenvalues $\lambda_i$ and orthonormal eigenvectors $u_i$, which encode structural properties.

\textbf{Cumulative Spectral Gradient (CSG) Computation:}  
Defines a complexity measure based on the differences between consecutive eigenvalues of the Laplacian, summing them cumulatively to assess how the graph’s structure evolves across its spectrum. Theoretically, the CSG quantifies the cumulative effect of spectral gaps, reflecting the progressive separation of classes and providing a nuanced view of complexity that ties directly to the graph’s global properties. This is particularly relevant for tail prediction, as it indicates the degree of variation in prediction difficulty across the dataset. The CSG measures complexity via eigenvalue differences. Order the eigenvalues:
\begin{equation}
    0 = \lambda_0 \le \lambda_1 \le \ldots \le \lambda_{K-1},
\end{equation}
\begin{equation}
    \text{Define gaps,} ~ \delta_i = \lambda_{i+1} - \lambda_i, \quad i = 0, 1, \ldots, K-2,
\end{equation}
\begin{equation}
    \text{Then,} ~\text{CSG}_{k_c} = \sum_{i=0}^{k_c - 1} \delta_i = \lambda_{k_c} - \lambda_0,
\end{equation}
\begin{equation}
   \text{and,} ~  \text{CSG}_{K-1} = \lambda_{K-1} - \lambda_0.
\end{equation}
\citet{branchaud2019spectral} claim that higher CSG values indicate higher complexity (more class overlap); lower CSG values indicate better separation and easier tail prediction.
\subsection{Experiments}
\textbf{Datasets:} The following datasets are used: \textbf{FB15k-237} \cite{bollacker2008freebase} consists of 14,541 entities, 237 relations, and a total of 310,116 triplets. \textbf{WN18RR} \cite{miller1995wordnet} includes 40,943 entities, 11 relations, and a total of 92,583 triplets. \textbf{CoDEx-S} \cite{safavi2020codex} features 2,034 entities, 42 relations, and a total of 40,198 triplets. \textbf{CoDEx-M} \cite{safavi2020codex} has 17,050 entities, 51 relations, and a total of 185,584 triplets. \textbf{CoDEx-L} \cite{safavi2020codex} includes 77,951 entities, 69 relations, and a total of 673,872 triplets. \textbf{Countries} \cite{10577554} dataset contains 271 entities, 2 relations, and 1159 triplets. \textbf{Toy} \cite{10577554} includes 278 entities, 112 relations, and 4826 triplets. \textbf{UML} \cite{Bodenreider_2004} contains 135 entities, 46 relations, and 6529 triplets. \textbf{Nations} \cite{10577554} has 14 entities, 55 relations, and 1992 triplets. All the datasets are publicly available online \footnote{https://tinyurl.com/mr8ckwmb}.

\subsection{Results}
\begin{figure}[h!]
\begin{center}
\includegraphics[width=\columnwidth, height=0.3\textheight]{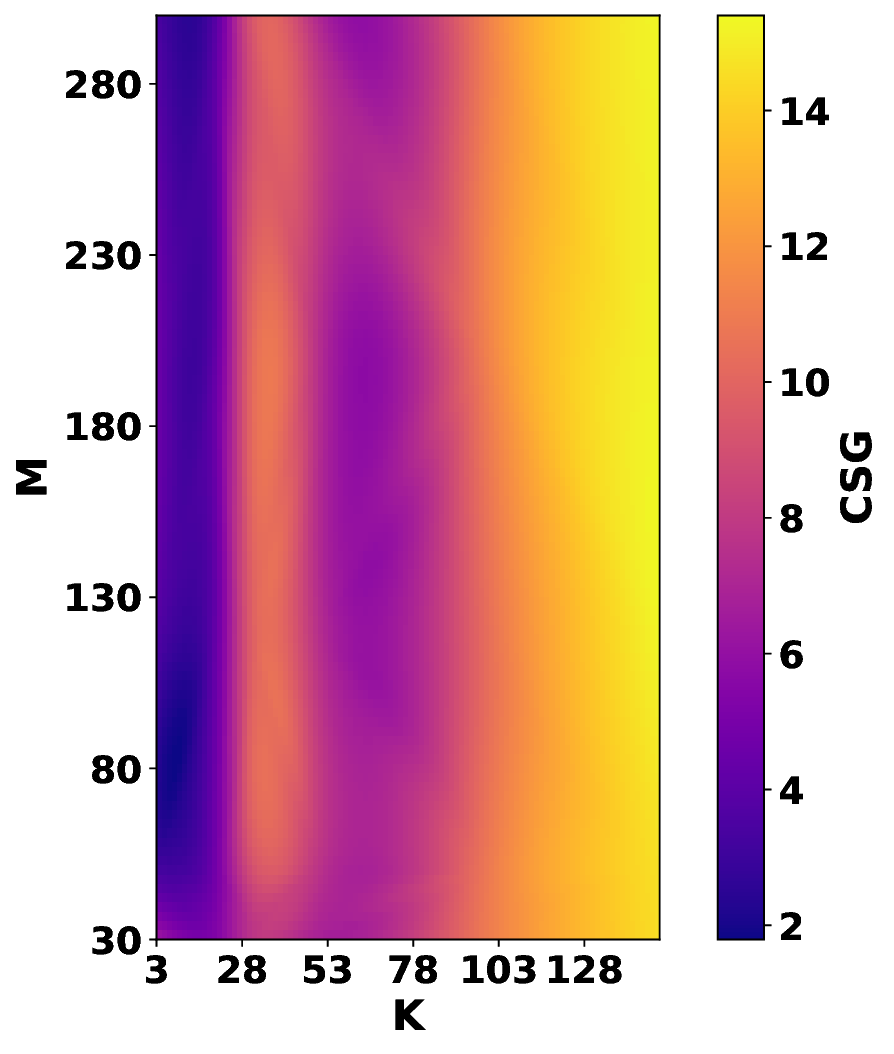}
\caption{\scriptsize CSG as a function of $M$ and $K$ values.}
\label{k-s-codex}
\end{center}
\end{figure}

Figure \ref{k-s-codex} illustrates the Cumulative Spectral Gradient (CSG) of Codex-S dataset in the tail-prediction task setting, represented as a surface function of parameters $K$ and $M$. Both parameters influence the CSG values, the impact of $K$ on CSG however is quite significant. This observation counters the previously held belief, as discussed in \cite{branchaud2019spectral}, where it was argued that these parameters had minimal influence. In contrast, our findings provide compelling evidence that both $K$ and $M$ play a critical role in shaping the spectral complexity assessment. Specifically, increasing $K$ leads to higher CSG values (in general), reflecting an increased perception of dataset complexity. It is likely that, smaller $K$ values tend to ignore large scale structural features (like connectivity patterns) within the data, thereby missing to capture fine-grained variations that capture complex interactions and class overlap. This results in a lower perceived complexity. The interaction between $K$ and $M$ also reveals key insights about their joint influence on complexity estimation. Specifically, for smaller $K$ values, $M$ plays a critical stabilizing role, as low $K$ is highly sensitive to sampling effects. 

\begin{figure}[ht]
\begin{center}
\centerline{\includegraphics[width=1.0 \columnwidth]{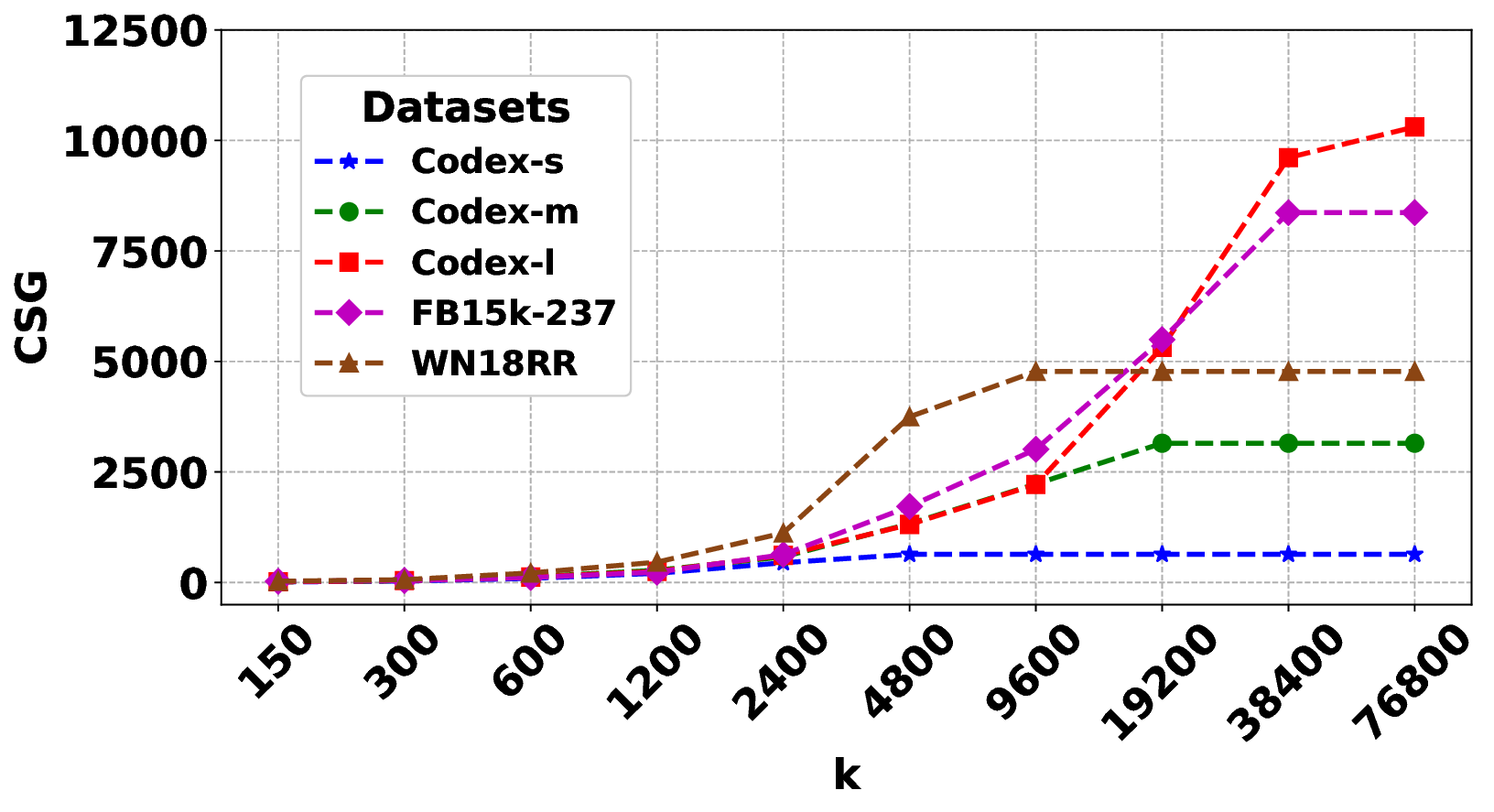}} 
\caption{\scriptsize A plot of CSG as a function of $K$ values at $M = 100$.}
\label{k_csg}
\end{center}
\end{figure}

Additionally, Figure \ref{k_csg} illustrates how the CSG is strongly influenced by the parameter $K$, with CSG values increasing consistently as $K$ increases across all datasets. This trend reveals that larger $K$-values capture broader structural patterns, leading to higher perceived complexity, while smaller $K$-values emphasize local structure and result in lower CSG.
Furthermore, the variation in CSG across different datasets highlights the importance of tailoring $K$ to the specific structural and semantic characteristics of each KG.

\begin{figure}[ht]
\begin{center}
\centerline{\includegraphics[width=1.1\columnwidth]{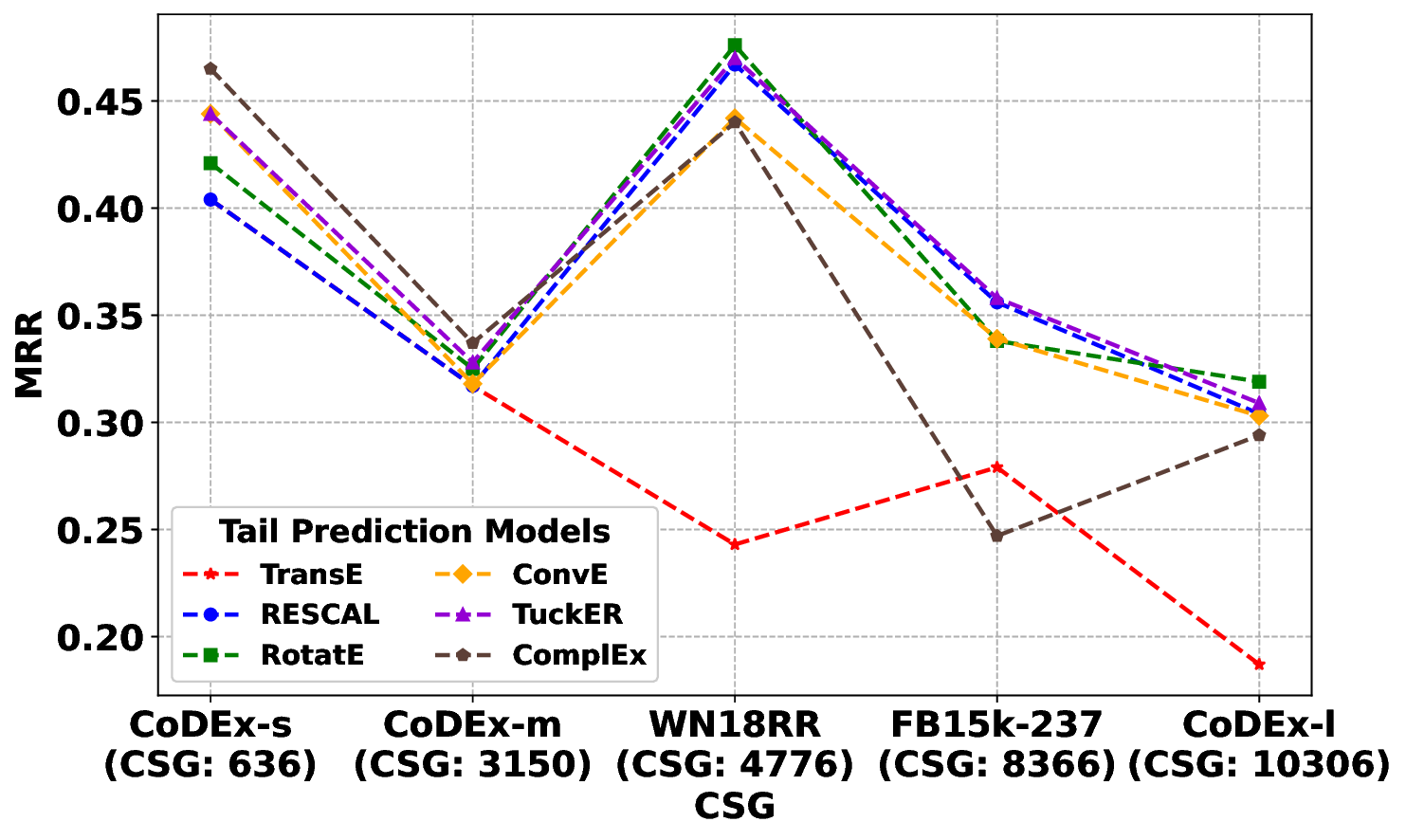}} 
\caption{\scriptsize Relationship Between MRR from different tail-prediction models on five standard KG datasets and the corresponding CSG values.}
\label{k-s-csg-emb}
\end{center}
\end{figure}

Finally, Figure~\ref{k-s-csg-emb} plots CSG values for five standard KG benchmarks against the corresponding MRR values achieved by a suite of tail‐prediction models. Contrary to \citet{branchaud2019spectral}, we observe no meaningful correlation (mean Pearson coefficient \(R = - 0.644\)) between CSG and model performance across all datasets and methods. In summary, contrary to the assertion that CSG can consistently forecast downstream performance, these findings cast doubt on its value as a model‐independent separability measure for large‐scale classification tasks specially in KG domain and underscore the necessity for more reliable metrics in KG evaluation. 

\vspace{-5pt}
\section{Conclusion}
CSG is significantly influenced by parameters, $K$ and $M$, challenging prior assumptions of their minimal impact on complexity assessments as well as the application of CSG as a reliable complexity metric for large multi-class datasets.

\bibliography{example_paper}
\bibliographystyle{icml2025}

\end{document}